\documentclass[letterpaper,10pt,conference]{ieeeconf}
\IEEEoverridecommandlockouts
\usepackage{amsmath,amssymb,bm}
\usepackage{booktabs}
\usepackage{multirow}
\usepackage{graphicx}
\usepackage{algorithm}
\usepackage{algpseudocode}
\usepackage{cite}
\usepackage{xcolor}
\usepackage{placeins}
\usepackage{array}
\usepackage{booktabs}
\usepackage{tabularx}
\usepackage{booktabs,tabularx}
\usepackage{url}

\newcolumntype{L}[1]{%
  >{\raggedright\arraybackslash}p{#1}%
}

\title{\LARGE \bf
CorrRisk-WM: Corridor-Conditioned Risk World Modeling for Safety-Critical Trajectory Planning}

\author{Tingyu Guo and Reza Langari%
\thanks{Tingyu Guo and Reza Langari are with the
J. Mike Walker '66 Department of Mechanical Engineering,
Texas A\&M University, College Station, TX 77843, USA.}}

\begin{document}
\maketitle
\thispagestyle{empty}
\pagestyle{empty}

\begin{abstract}
Safe local planning requires forecasting surrounding-agent motion
and evaluating candidate-specific risks, since identical agent
motion can pose different risks to different ego trajectories.
We present CorrRisk-WM, a planning-oriented partial world model
coupling environment evolution with supervised intrusion and
near-miss prediction over bounded candidate-trajectory corridors.
A latent environment model recursively predicts agent states
and updates agent--agent and agent--map interactions.
Each candidate queries the evolving environment through
footprint-aware geometry and learned agent--corridor representations.
A lightweight recurrent risk module uses temporal context to
estimate per-slice hazards; survival aggregation yields first-entry
and horizon-level event probabilities.
On $29{,}176$ scenarios from 100 Waymo validation shards,
CorrRisk-WM achieves intrusion average precision (AP) of $0.8567$ and $1$-m near-miss
first-entry AP of $0.8671$. In baseline comparisons, it attains
the highest near-miss AP at all three distance thresholds and
the lowest observed open-loop collision rate ($4.88\%$), with
route progress of $15.35$~m.
Across three seeds, removing dynamic environment modeling or
candidate-conditioned geometric interaction reduces mean intrusion
AP from $0.8590$ to $0.7624$ and $0.7252$, respectively.
These results support coupling environment evolution with
candidate-conditioned geometric reasoning for risk prediction
and safety-oriented candidate selection.
\end{abstract}

\section{Introduction}
Safe local planning requires generating dynamically feasible
trajectories while anticipating surrounding traffic over the
planning horizon. Existing approaches address these requirements
through interaction-aware behavior planning~\cite{epsilon},
hierarchical trajectory generation~\cite{hierarchical_trajectory},
and constrained trajectory optimization~\cite{cfs}.

Interactive forecasting, occupancy-based planning, and driving world
models support future prediction and trajectory
evaluation~\cite{gameformer,quad,world4drive,wote}.
Foundation-model research further connects multimodal scene
understanding with reasoning and decision
making~\cite{fm_av_review}. 

For a fixed agent-motion forecast, candidate trajectories may
differ in overlap, clearance, and encounter timing, making risk
candidate-dependent rather than a property of agent states alone.
Corridors link free-space geometry, predicted risk, and trajectory
optimization~\cite{iris,rast,driveincorridors,corridor_assessment}.
We introduce \textbf{CorrRisk-WM}, which treats each candidate
corridor as a bounded spatiotemporal query over an evolving local
environment. Intrusion and clearance-based near-miss probabilities,
together with first-entry timing, give candidate-specific risk
an explicit geometric interpretation. Under a non-reactive
assumption, one logged scene future provides event supervision
for multiple ego candidates.
Our main contributions are as follows:
\begin{itemize}
    \item We formulate candidate-specific risk as structured
    agent--candidate--time corridor-event prediction, with explicit
    intrusion and multi-threshold near-miss targets.

    \item We develop \emph{Dynamic Environment Modeling}, combining
    latent transitions with a supervised state probe to evolve agent states and refresh agent--agent (AA) and agent--map (AM) context.

    \item We introduce \emph{Candidate-Conditioned Geometric
    Interaction Modeling}, integrating relevant-agent routing,
    footprint-aware geometry, and learned agent--corridor fusion.
    \emph{Recurrent Temporal Risk Modeling} carries candidate-specific
    risk context across time slices to support structured risk
    estimation.

    \item We systematically evaluate the framework on the
    Waymo Open Motion Dataset~\cite{waymo_motion}, assessing risk
    prediction and candidate selection through baseline comparisons
    and examining the role of each individual modeling component
    through ablation studies.
\end{itemize}
\begin{figure}[t]
    \centering
    \setlength{\abovecaptionskip}{2pt}
    \setlength{\belowcaptionskip}{0pt}
    \includegraphics[width=1.0\linewidth]{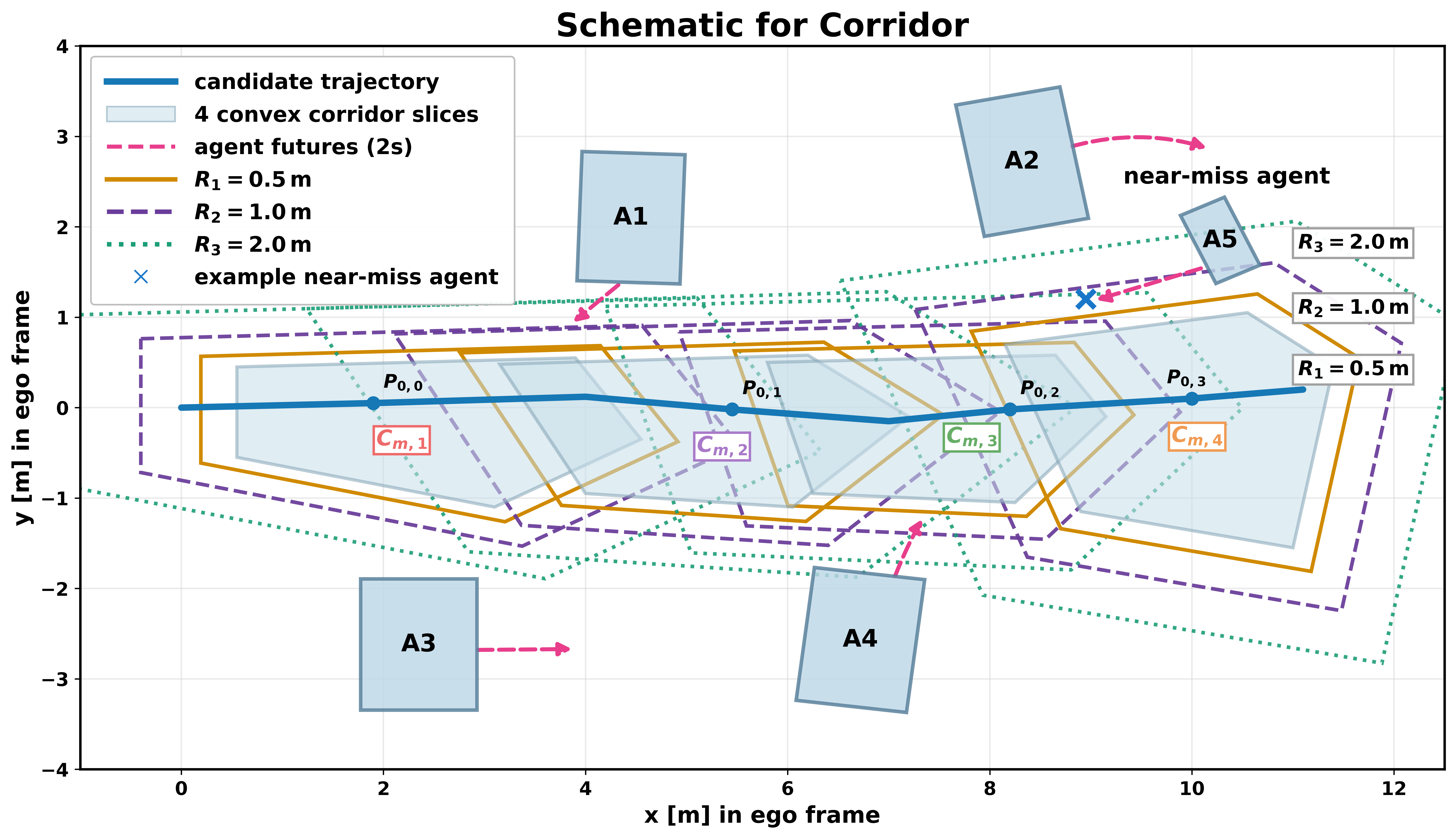}
    \caption{Temporal corridor representation: a candidate trajectory
    (blue), four shaded corridor slices, near-miss contours at $0.5$,
    $1.0$, and $2.0\,\mathrm{m}$, and surrounding agents A1--A5 with
    $2$-s future motion.}
    \label{fig:corridor}
\end{figure}

\section{Related Work}

\subsection{Interactive Prediction and Candidate Evaluation}

MotionDiffuser models multimodal joint futures through
diffusion~\cite{motiondiffuser}, conditional behavior prediction forecasts other agents' trajectories conditioned on a queried ego
trajectory~\cite{conditional_behavior}, and GameFormer uses
hierarchical game-theoretic reasoning for interactive prediction
and planning~\cite{gameformer}. GenAD and DiffusionDrive further
couple scene representations with generative trajectory
planning~\cite{genad,diffusiondrive}.

QuAD evaluates ego trajectories through occupancy queries at
planning-relevant spatiotemporal locations~\cite{quad}.
NAVSIM instead provides a scalable non-reactive simulation
benchmark for evaluating planning policies using logged
data~\cite{navsim}. 

\subsection{Driving World Models}

World models learn predictive internal dynamics for planning
and control~\cite{worldmodels,dreamerv3,tdmpc2}.
Surveys review applications in future generation, behavior planning,
and prediction--planning
interaction~\cite{initial_wm_survey,wmad_survey}.
Driving applications include visual future generation
~\cite{gaia1,drivewm,drivedreamer},
occupancy forecasting~\cite{occworld},
4D scene-representation pretraining~\cite{driveworld},
and joint future image--depth prediction~\cite{unifuture}.

TrafficBots uses a scene-centric vectorized representation for
multi-agent simulation and motion prediction~\cite{trafficbots}. CarDreamer provides configurable driving tasks and integrated world-model backbones for developing
and evaluating world-model-based control~\cite{cardreamer}.

World4Drive generates intention-conditioned candidate trajectories
and evaluates them using predicted latent futures~\cite{world4drive},
while WoTE predicts future bird's-eye-view (BEV) states for
online trajectory evaluation~\cite{wote}. RiskWorld predicts object-level risk scores and temporal risk
curves from rolled-out ego--object relations, deriving
time-to-risk by threshold crossing~\cite{riskworld}.
CorrRisk-WM instead uses ego-candidate corridor queries and discrete-time hazards to predict agent-wise first-occurrence distributions for intrusion and multi-threshold near misses.

\subsection{Corridor-Based and Semantic Risk Representations}

IRIS and safe flight corridors exploit obstacle-free convex
regions for trajectory optimization~\cite{iris,sfc}, while
convex feasible-set methods address nonconvex planning
constraints through iterative convexification~\cite{cfs}.
RAST constructs risk-aware spatiotemporal corridors for
micro aerial vehicle navigation using regional collision risk
derived from predicted dynamic maps~\cite{rast}.

Drive in Corridors learns intermediate corridor constraints for
trajectory optimization~\cite{driveincorridors}, while
transformer-based corridor assessment predicts drivability and
expected trajectory cost~\cite{corridor_assessment}.

CorrRisk-WM uses candidate corridors as agent-specific
event-query regions, complementing corridor construction
and downstream trajectory optimization.

%

\section{Problem Formulation}
\label{sec:problem}

We formulate candidate evaluation in an ego-centric planar frame aligned
with the current ego heading. All states, map features, candidates, and
corridor geometry are expressed in this frame.

\subsection{Scene Representation and Candidate Trajectories}
\label{sec:problem_inputs}

For surrounding agent $i$ at discrete time index $t$, with $t=0$
denoting the current planning instant and $t<0$ denoting past
observations, the state is
\begin{align}
\bm x_{i,t}
&=
[\bm p_{i,t};
 \bm d^{\psi}_{i,t};
 \bm v_{i,t};
 \ell_i;
 w_i]
\in\mathbb R^8,\\
\bm d^{\psi}_{i,t}
&=
[\cos\psi_{i,t},\sin\psi_{i,t}]^{\top},
\label{eq:agent_state}
\end{align}
where $\bm p_{i,t}\in\mathbb R^2$, $\bm v_{i,t}\in\mathbb R^2$,
$\psi_{i,t}$, and $(\ell_i,w_i)$ denote position, velocity, heading,
and footprint length and width, respectively. The observed motion history and categorical type of agent $i$ are
\begin{equation}
\bm H_i
=
\bigl(\bm x_{i,t}\bigr)_{t=-(N_{\rm hist}-1)}^{0},
\qquad
\kappa_i\in\mathcal A_{\rm type},
\label{eq:agent_history}
\end{equation}

The local map comprises vectorized Waymo roadgraph polylines within the
ego-centric window
$\mathcal B$:
\begin{equation}
\mathcal M=\{\mathcal P_p\}_{p=1}^{P},
\qquad
\mathcal P_p=\{\bm m_{p,r}\}_{r=1}^{L_p},
\label{eq:map_polyline}
\end{equation}
where $p$ and $r$ index polylines and their valid samples, respectively. Each roadgraph sample is represented as
\begin{equation}
\bm m_{p,r}
=
[x_{p,r},y_{p,r},
 d^{x}_{p,r},d^{y}_{p,r}]^{\top}
\in\mathbb R^4,
\label{eq:map_point}
\end{equation}
where $(x_{p,r},y_{p,r})$ and
$(d^{x}_{p,r},d^{y}_{p,r})$ denote the position and local roadgraph direction of sample \(r\) on polyline \(p\).
We retain at most $P$ polylines and $L_p\leq L_{max}$ valid samples per
polyline.

A shared upstream sampler produces $M$ ego candidates :
\begin{equation}
\mathcal T=\{\tau^m\}_{m=1}^{M},
\qquad
\tau^m=\{\bm s^{\rm SDC}_{m,\nu}\}_{\nu=1}^{L_{max}},
\label{eq:candidate_set}
\end{equation}
where $m$ indexes ego candidates and $\nu$ indexes sampled future states along candidate $m$. The corresponding ego state is
\begin{equation}
\bm s^{\rm SDC}_{m,\nu}
=
[\bm p^{\rm SDC}_{m,\nu};
 \bm d^{\psi,\rm SDC}_{m,\nu};
 \bm v^{\rm SDC}_{m,\nu}]
\in\mathbb R^6,
\label{eq:candidate_state}
\end{equation}

\subsection{From Candidate Trajectories to IRIS Corridors}
\label{sec:problem_corridor}
Each candidate comprises $K$ intervals
$[t_{m,k}^{\mathrm{start}},\,t_{m,k}^{\mathrm{end}}]$
of duration $\Delta$.
Each interval has $J$ substeps of duration $\delta=\Delta/J$,
with total horizon $H_T=K\Delta$. and $\nu=kJ+j$, $S^{\mathrm{seed}}_{m,k}$ is the IRIS seed region centered and
aligned with $\bm s^{\mathrm{SDC}}_{m,k}$. The deterministic chain is
\begin{equation}
\left.\tau^m\right|_{
[t_{m,k}^{\rm start},\,t_{m,k}^{\rm end}]}
\longrightarrow
\bm s^{\rm SDC}_{m,k}
\longrightarrow
S^{\rm seed}_{m,k}
\xrightarrow{\operatorname{IRIS}(\,\cdot\,; o_{m,k})}
C_{m,k}.
\label{eq:trajectory_to_corridor}
\end{equation}
$o_{m,k}$ comprises road boundaries and inflated agent footprints predicted at the $k$-th of $K$ slice midpoints under constant velocity. IRIS alternates obstacle-separating halfspace construction and maximum-volume ellipsoid inflation to form an a priori geometric space~\cite{iris}, producing
\begin{align}
C_{m,k}
&=\{\bm p_c\in\mathbb R^2:\bm H^C_{m,k}\bm p_c\leq\bm b^C_{m,k}\},
\label{eq:iris_slice}\\
\mathcal C_m
&=\{(C_{m,k},[t_{m,k}^{\mathrm{start}},\,t_{m,k}^{\mathrm{end}}])\}_{k=0}^{K-1}.
\label{eq:temporal_corridor}
\end{align}
Thus $\bm H^C_{m,k}$ and $\bm b^C_{m,k}$ define the convex polygonal slice, $\bm p_c$ denotes an arbitrary 2-D point in the ego-centric frame.

The exact IRIS polygon is stored by its vertices:
\begin{equation}
\mathcal V^C_{m,k}
=
\{v^{(q)}_{m,k}\}_{q=1}^{Q_{m,k}},
\quad
v^{(q)}_{m,k}\in\mathbb R^2,
\quad
C_{m,k}
=
\operatorname{conv}(\mathcal V^C_{m,k}),
\label{eq:corridor_vertices}
\end{equation}
where $Q_{m,k}$ is the number of valid vertices.

Each slice is cached as a compact descriptor of $\mathcal V^C_{m,k}$, its temporal interval, and a prior-risk feature:

\begin{equation}
\begin{split}
\bm\eta_{m,k}=[&c^x_{m,k},c^y_{m,k},
\cos\theta_{m,k},\sin\theta_{m,k},
 h^w_{m,k},h^\ell_{m,k},\\[-1mm]
& t^{start}_{m,k},t^{end}_{m,k},\pi^{\rm prior}_{m,k}]^{\top}
\in\mathbb R^9.
\end{split}
\label{eq:corridor_descriptor}
\end{equation}
Here $(c^x_{m,k},c^y_{m,k})$ is the slice center,
$\theta_{m,k}$ is its longitudinal orientation,
$h^w_{m,k}$ and $h^\ell_{m,k}$ are the half-width and half-length of its
oriented rectangular summary, $\pi^{\rm prior}_{m,k}$ denotes prior risk.

\subsection{Candidate-Conditioned Risk Formulation}

For candidate $m$, let $\mathcal I_m^{\rm reach}$ denote the set of
valid agents that can physically approach its temporal corridor.
Define
\[
\bar t_{m,k}
=
\frac{t_{m,k}^{\rm start}+t_{m,k}^{\rm end}}{2},
\qquad
N_m
=
\min\!\left\{N,\left|\mathcal I_m^{\rm reach}\right|\right\}.
\]
The candidate-local agent pool is
\begin{equation}
\mathcal I_m
=
\underset{
\substack{
\mathcal U\subseteq\mathcal I_m^{\rm reach}\\
|\mathcal U|=N_m
}}
{\arg\min}
\sum_{i\in\mathcal U}
\min_{0\leq k<K}
d_{\rm sel}\!\left(
\bm p_{i,0}+\bm v_{i,0}\bar t_{m,k},
C_{m,k}
\right),
\label{eq:candidate_local_pool}
\end{equation}
where $d_{\rm sel}(\cdot,C_{m,k})$ denotes the point-to-corridor
ranking distance used after reachability filtering. Pools containing
fewer than $N$ agents are zero-padded and masked.

The resulting candidate context is
\begin{equation}
\mathcal O_m
=
\left(
\{(\bm x_{i,0},\bm H_i,\kappa_i):i\in\mathcal I_m\},
\mathcal M
\right).
\label{eq:candidate_context}
\end{equation}

At slice $k$, deterministic geometry selects
$\mathcal A_{m,k}\subseteq\mathcal I_m$ with
$|\mathcal A_{m,k}|\leq N_{\mathrm{sel}}$.
For each active agent, 
\begin{equation}
\widehat{\bm Y}_{i,m,k}
=
\mathcal F_\theta(\mathcal O_m,\mathcal C_m, \tau^m)_{i,k},
\quad i\in\mathcal A_{m,k}.
\label{eq:risk_query}
\end{equation}
where $\widehat{\bm Y}_{i,m,k}$ represents decoded risk package. $\mathcal F_{\theta}$ denotes the end-to-end CorrRisk-WM
mapping.

\section{Method}
\label{sec:method}
\begin{figure*}[t]
    \centering
    \includegraphics[
        width=1.0\textwidth,
        trim={0mm 45mm 0mm 25mm},
        clip
    ]{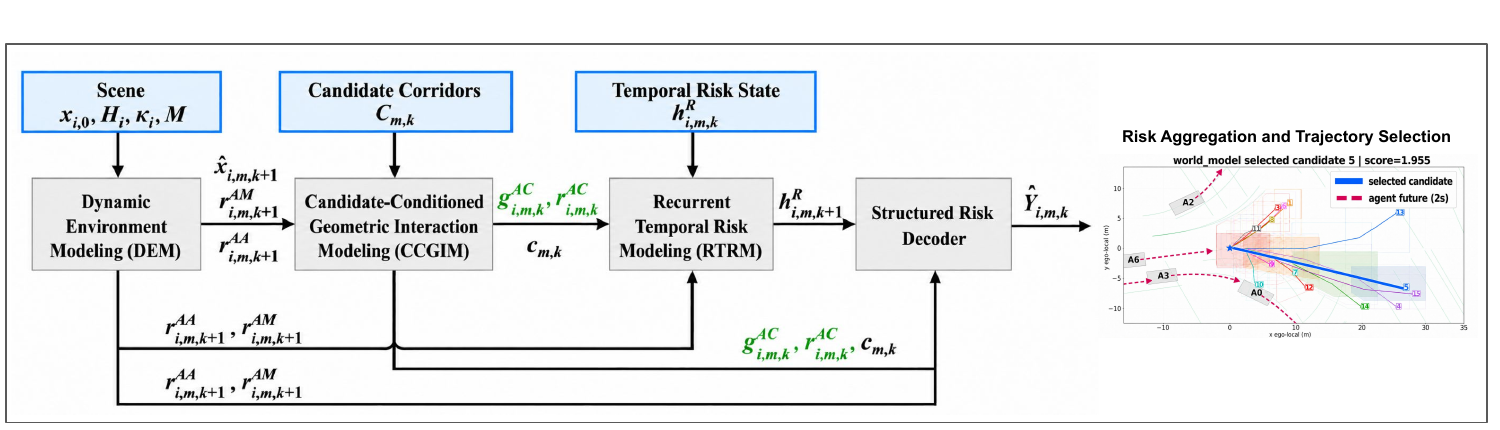}
    \caption{Overview of the CorrRisk-WM framework.}
    \label{fig:framework}
\end{figure*}

\subsection{Dynamic Environment Modeling (DEM)}
\label{sec:environment}
Each candidate $m$ is rolled out independently with its $\mathcal I_m$ and supervised using the same logged future.
\paragraph{Scene encoding}
The initial agent latent, candidate-local rollout state, and map tokens are
\begin{equation}
\bm z^{\rm E}_{i,m,0}
=
E_{\rm E}(\bm x_{i,m, 0},\bm H_i,\kappa_i),\\
\widehat{\bm x}_{i,m,0}=\bm x_{i, m, 0},\\
\bm\mu_p=E_{\rm M}(\mathcal P_p),
\label{eq:initial_environment}
\end{equation}
where $E_{\rm E}$ and $E_{\rm M}$ are the environment and map encoders,
respectively, and $\bm{\mu}_p\in\mathbb{R}^{d}$ is the token of polyline $p$, $d$ denotes the latent feature dimension. 

\paragraph{Dynamic relational context}
Let $\widehat{\bm p}_{i, m, k}$ and $\widehat{\bm v}_{i, m, k}$ denote the
position and velocity of agent $i$ at slice $k$. For agents $i$ and $n$,
\begingroup
\medmuskip=2mu
\thickmuskip=1.5mu
\begin{gather}
\Delta\bm p_{in,m,k}
=\widehat{\bm p}_{i,m,k}-\widehat{\bm p}_{n,m,k},
\;
\Delta\bm v_{in,m,k}
=\widehat{\bm v}_{i,m,k}-\widehat{\bm v}_{n,m,k},
\nonumber\\
d_{in,m,k}
=\|\Delta\bm p_{in,m,k}\|_2,
\;
v^{\rm close}_{in,m,k}
=-\frac{
\Delta\bm p_{in,m,k}^{\top}\Delta\bm v_{in,m,k}
}{
d_{in,m,k}+\varepsilon_{close}
}.
\label{eq:aa_geometry}
\end{gather}
\endgroup
The agent--agent descriptor and bias are
\begin{equation}
\begin{aligned}
\bm\phi^{\rm AA}_{in,m,k}
=
\bigl[
&\Delta\bm p_{in,m,k};
 \Delta\bm v_{in,m,k};
 d_{in,m,k};
 \|\Delta\bm v_{in,m,k}\|_2;\\[-1mm]
&(d_{in,m,k}+1)^{-1};
 v^{\rm close}_{in,m,k}
\bigr]
\in\mathbb R^8 .
\end{aligned}
\label{eq:aa_descriptor}
\end{equation}

\begin{equation}
\left[\bm B^{\rm AA}_{m,k}\right]_{in}
=
b_{\rm AA}\!\left(\bm\phi^{\rm AA}_{in,m,k}\right).
\label{eq:aa_bias}
\end{equation}
For polyline $p$, let $\bar{\bm q}_p$ and $\bar{\bm t}_p$ denote its
mean position and normalized mean direction, respectively, with
$\bar{\bm n}_p=[-\bar t_p^y,\bar t_p^x]^\top$. Define
\begin{equation}
\begin{gathered}
\Delta\bm p_{ip,m,k}
=
\widehat{\bm p}_{i,m,k}-\bar{\bm q}_p,
\qquad
d_{ip,m,k}
=
\|\Delta\bm p_{ip,m,k}\|_2,
\\[-1mm]
\bm u^v_{i,m,k}
=
\frac{\widehat{\bm v}_{i,m,k}}
{\max\!\left\{
\|\widehat{\bm v}_{i,m,k}\|_2,
\varepsilon_{\rm norm}
\right\}} .
\end{gathered}
\label{eq:am_geometry}
\end{equation}
The agent--map descriptor and bias are
\begin{equation}
\begin{aligned}
\bm\phi^{\rm AM}_{ip,m,k}
=
\bigl[
&\Delta\bm p_{ip,m,k};
 d_{ip,m,k};
 \Delta\bm p_{ip,m,k}^{\top}\bar{\bm n}_p;
 \Delta\bm p_{ip,m,k}^{\top}\bar{\bm t}_p;\\[-1mm]
&\|\widehat{\bm v}_{i,m,k}\|_2;
 (\bm u^v_{i,m,k})^\top\bar{\bm t}_p;
 (d_{ip,m,k}+1)^{-1}
\bigr]
\in\mathbb R^8 .
\end{aligned}
\label{eq:am_descriptor}
\end{equation}

\begin{equation}
\left[\bm B^{\rm AM}_{m,k}\right]_{ip}
=
b_{\rm AM}\!\left(\bm\phi^{\rm AM}_{ip,m,k}\right).
\label{eq:am_bias}
\end{equation}
where $b_{\rm AA}$, $b_{\rm AM}$ are scalar MLPs (Multilayer Perceptron).

Let
\begin{equation}
\bm Z_{m,k}^{\rm E}
=
[\bm z^{\rm E}_{i, m, k}]_{i\in\mathcal I_m},
\qquad
\bm Z^{\rm M}
=
[\bm\mu_p]_{p=1}^{P},
\end{equation}
The geometry-biased relational contexts are
\begin{align}
\bm R_{m,k}^{\rm AA}
&=
\operatorname{AA}
(\bm Z_{m,k}^{\rm E};\bm B_{m,k}^{\rm AA}),
\\
\bm R_{m,k}^{\rm AM}
&=
\operatorname{AM}
(\bm R_{m,k}^{\rm AA},\bm Z^{\rm M};\bm B_{m,k}^{\rm AM}),
\label{eq:dynamic_relations}
\end{align}
where
$\bm B_{m,k}^{\rm AA}\in\mathbb R^{N\times N}$ and
$\bm B_{m,k}^{\rm AM}\in\mathbb R^{N\times P}$ are additive attention
biases. The agent-wise outputs are $\bm r^{\rm AA}_{i,m,k}=[\bm R^{\rm AA}_{m,k}]_{i,:}$ and $\bm r^{\rm AM}_{i,m,k}=[\bm R^{\rm AM}_{m,k}]_{i,:}$; $\operatorname{AA}$ and $\operatorname{AM}$ denote the self-attention and cross-attention blocks, respectively.

\paragraph{Latent transition and future-anchored state probe}
The corridor-unconditioned environment transition is
\begin{equation}
\bm z^{\mathrm E}_{i, m, k+1}
=T_{\mathrm E}(\bm z^{\mathrm E}_{i, m, k},
\bm r^{\mathrm{AA}}_{i, m, k},\bm r^{\mathrm{AM}}_{i, m, k}),
\label{eq:environment_transition}
\end{equation}
where $T_{\rm E}$ is the learned environment transition. $D_X$ decodes $J$ substeps:
\begin{multline}
\left\{
\left(
\Delta\widehat{\bm p}_{i, m, k,j},
\widetilde{\bm d}^{\psi}_{i, m, k, j},
\widehat{\bm v}_{i, m, k, j},
\widehat{\bm a}_{i, m, k, j},
\widehat{\omega}_{i, m, k, j}
\right)
\right\}_{j=1}^{J}\\
=
D_X\!\left(
\bm z^{\mathrm E}_{i, m, k},
\bm z^{\mathrm E}_{i, m, k+1},
\widehat{\bm x}_{i, m, k}
\right).
\label{eq:state_probe}
\end{multline}
Here, $\widehat{\bm a}_{i,m,k,j}\in\mathbb R^2$ and
$\widehat\omega_{i,m,k,j}\in\mathbb R$ denote the acceleration
and yaw rate predicted by $D_X$, respectively. The substep states
are reconstructed as
\begin{equation}
\begin{gathered}
\widehat{\bm d}^{\psi}_{i,m,k,j}
=
\frac{\widetilde{\bm d}^{\psi}_{i,m,k,j}}
{\max\!\left\{
\|\widetilde{\bm d}^{\psi}_{i,m,k,j}\|_2,
\varepsilon_{\rm norm}
\right\}},
\\
\widehat{\bm p}_{i,m,k,j}
=
\widehat{\bm p}_{i,m,k}
+\Delta\widehat{\bm p}_{i,m,k,j},
\\
\widehat{\bm x}_{i,m,k,j}
=
[\widehat{\bm p}_{i,m,k,j};
 \widehat{\bm d}^{\psi}_{i,m,k,j};
 \widehat{\bm v}_{i,m,k,j};
 \ell_i;w_i],
\\
\widehat{\bm x}_{i,m,k+1}
=
\widehat{\bm x}_{i,m,k,J}.
\end{gathered}
\label{eq:state_reconstruction}
\end{equation}
Here, $\Delta\widehat{\bm p}_{i,m,k,j}$ is measured from the
start of slice $k$. 

Applying Eqs.~(16)--(24) to $\widehat{\bm x}_{i,m,k+1}$
gives the updated agent-wise contexts
$\bm r^{\mathrm{AA}}_{i,m,k+1}$ and
$\bm r^{\mathrm{AM}}_{i,m,k+1}$.

\subsection{Candidate-Conditioned Geometric Interaction Modeling (CCGIM)}
\label{sec:ac_interaction}
For candidate $m$, the ordered temporal corridor slices are encoded as
\begin{equation}
\{\bm c_{m,k}\}_{k=0}^{K-1}
=E_{\mathrm C}\!\left(
\{\bm\eta_{m,k},\bm s^{\rm SDC}_{m,k},\mathcal V^C_{m,k}\}_{k=0}^{K-1}
\right).
\label{eq:corridor_tokens}
\end{equation}
Here, $E_{\rm C}$ embeds the slice attributes through self-attention.

For predicted substep $(i,m, k,j)$, define the heading and oriented agent
footprint by
\begin{equation}
\widehat\psi_{i,m,k,j}
=
\operatorname{atan2}\!\left(
\widehat d^{\psi,y}_{i,m,k,j},
\widehat d^{\psi,x}_{i,m,k,j}
\right).
\label{eq:predicted_heading}
\end{equation}

\begin{equation}
\widehat F_{i,m,k,j}
=
\operatorname{Rect}\!\left(
\widehat{\bm p}_{i,m,k,j},
\widehat\psi_{i,m,k,j},
\ell_i,w_i
\right).
\label{eq:predicted_footprint}
\end{equation}
where $\operatorname{Rect}$ denotes an oriented-rectangle
construction operator.
The footprint-aware clearance is
\begin{equation}
d^{\rm foot}_{i,m,k,j}=d_{\pm}(\widehat F_{i, m, k,j},C_{m,k}).
\label{eq:predicted_clearance}
\end{equation}
Using polygon scale \(s^C_{m,k}\), the normalized margin channels are
\begin{equation}
\bm\delta^C_{i,m,k,j}
=
\operatorname{clip}\!\left(
-\frac{d^{\rm foot}_{i,m,k,j}}{s^C_{m,k}},
-b_\delta,b_\delta
\right)\bm 1_2 .
\label{eq:corridor_margins}
\end{equation}
The agent velocity in the corridor frame is
\begin{equation}
\begin{gathered}
\begin{bmatrix}
v^{\parallel}_{i,m,k,j}\\
v^{\perp}_{i,m,k,j}
\end{bmatrix}
=
\begin{bmatrix}
\cos\theta_{m,k} & \sin\theta_{m,k}\\
-\sin\theta_{m,k} & \cos\theta_{m,k}
\end{bmatrix}
\widehat{\bm v}_{i,m,k,j}.
\end{gathered}
\label{eq:corridor_velocity}
\end{equation}
With $\bm c^{xy}_{m,k}=[c^x_{m,k},c^y_{m,k}]^{\top}$ and
$\bm r^{xy}_{i,m,k,j}=\widehat{\bm p}_{i, m, k,j}-\bm c^{xy}_{m,k}$, the
approach speed and approximate time-to-corridor are
\begin{align}
v^{\rm app}_{i,m,k,j}
&=-\widehat{\bm v}_{i, m, k,j}^{\top}
\frac{\bm r^{xy}_{i,m,k,j}}
     {\|\bm r^{xy}_{i,m,k,j}\|_2+\varepsilon_{app}},\\
\mathrm{TTC}_{i,m,k,j}
&=
\operatorname{clip}\!\left(
\frac{[d^{\rm foot}_{i,m,k,j}]_+}
{\max(v^{\rm app}_{i,m,k,j},v_{\min})},
0,T_{\max}
\right).
\label{eq:approach_ttc}
\end{align}
where $[x]_+=\max(x,0)$. The complete ten-dimensional geometry descriptor is
\begin{equation}
\begin{split}
\bm\chi^{\mathrm{AC}}_{i,m,k,j}=[&d^{\rm foot},\bm\delta^C,
 v^{\parallel},v^{\perp},\|\widehat{\bm v}\|_2,\\[-1mm]
&v^{\rm app},\mathrm{TTC},s^{\rm size},\pi^{\rm prior}
]_{i,m,k,j}^{\top},
\end{split}
\label{eq:ac_descriptor}
\end{equation}
where $s^{\rm size}=(\ell_i+w_i)/2$

The minimum-clearance substep is encoded by \(E_g\) into the agent--corridor geometry embedding:
\begin{equation}
j^*_{i,m,k}=\arg\min_{1\leq j\leq J}d^{\rm foot}_{i,m,k,j},
\qquad
\bm g^{\mathrm{AC}}_{i,m,k}
=E_g(\bm\chi^{\mathrm{AC}}_{i,m,k,j^*_{i,m,k}}).
\label{eq:minimum_geometry}
\end{equation}
Quantities without a substep index below are evaluated at $j^*_{i,m,k}$. The
same geometry determines slice-wise relevance:
\begin{equation}
\begin{aligned}
s_{i,m,k}
={}&-d^{\rm foot}_{i,m,k}
+\beta_{\rm app}[v^{\rm app}_{i,m,k}]_{+} \\[-0.5mm]
&\hspace{-1em}
+\beta_{\rm ttc}
\exp\!\left(-\frac{\operatorname{TTC}_{i,m,k}}
{\tau_{\rm ttc}}\right)
+\beta_{\rm prox}
\mathbf 1\!\left[d^{\rm foot}_{i,m,k}\leq\rho_{\rm sel}\right].
\end{aligned}
\label{eq:relevance_score}
\end{equation}

\begin{equation}
\mathcal A_{m,k}
=
\operatorname{TopK}_{i\in\mathcal I_m}
\left(s_{i,m,k},N_{\rm sel}\right).
\label{eq:active_set}
\end{equation}
For an active agent, define the fusion input
\begin{equation}
\bm f^{\mathrm{AC}}_{i,m,k}
=[\bm z^{\mathrm E}_{i,m, k+1};\bm r^{\mathrm{AA}}_{i,m,k+1};
  \bm r^{\mathrm{AM}}_{i,m,k+1};\bm c_{m,k};
  \bm g^{\mathrm{AC}}_{i,m,k}].
\label{eq:ac_fusion_input}
\end{equation}
A learned gate produces
\begin{equation}
\begin{gathered}
\bm\gamma^{\mathrm{AC}}_{i,m,k}
=
\alpha_{i,m,k}\,
\sigma\!\left(
G_{\mathrm{AC}}(\bm f^{\mathrm{AC}}_{i,m,k})
\right),
\\
\bm r^{\mathrm{AC}}_{i,m,k}
=
\operatorname{LN}\!\left(
\bm r^{\mathrm{AM}}_{i,m,k+1}
+
\bm\gamma^{\mathrm{AC}}_{i,m,k}
\odot
V_{\mathrm{AC}}(\bm f^{\mathrm{AC}}_{i,m,k})
\right).
\end{gathered}
\label{eq:ac_fusion}
\end{equation}
where $\alpha_{i,m,k}$ masks selected, valid, reachable slots;
$G_{\mathrm{AC}}$ and $V_{\mathrm{AC}}$ are MLPs for gate logits
and candidate-conditioned residuals; $\sigma$ denotes the sigmoid function.

\subsection{Recurrent Temporal Risk Modeling (RTRM)}
\label{sec:risk_state}

Each valid candidate--agent slot is initialized by a shared learned vector,
$\bm h^{\mathrm R}_{i,m,0}=\bm h^{\mathrm R}_0$. At
slice $k$ is
\begin{equation}
\bm e^{\mathrm R}_{i,m,k}
=[\bm z^{\mathrm E}_{i,m, k+1};\bm r^{\mathrm{AA}}_{i, m, k+1};
  \bm r^{\mathrm{AM}}_{i, m, k+1};\bm r^{\mathrm{AC}}_{i,m,k};
  \bm c_{m,k};\bm g^{\mathrm{AC}}_{i,m,k}].
\label{eq:risk_evidence}
\end{equation}
The candidate-specific temporal state is updated by
\begin{equation}
\bm h^{\mathrm R}_{i,m,k+1}
=T_{\mathrm R}(\bm h^{\mathrm R}_{i,m,k},
               \bm e^{\mathrm R}_{i,m,k}),
\label{eq:risk_transition}
\end{equation}
where \(T_R\) is a recurrent transition;
\subsection{Structured Risk Decoder}
The decoder input is defined as follows:
\label{sec:risk_decoder}
\begin{equation}
\bm u^D_{i,m,k}
=
[\bm h^{\rm R}_{i,m,k+1};
 \bm c_{m,k};
 \bm g^{\rm AC}_{i,m,k};
 \bm r^{\rm AA}_{i,m,k+1};
 \bm r^{\rm AM}_{i,m,k+1};
 \bm r^{\rm AC}_{i,m,k}].
\label{eq:decoder_input}
\end{equation}

Independent task-specific prediction heads operate on the common feature
vector:
\begin{equation}
\widehat{\bm Y}_{i,m,k}
=
D_Y\!\left(\bm u^D_{i,m,k}\right).
\label{eq:risk_decoder}
\end{equation}
and form the decoded risk package
\begin{equation}
\begin{aligned}
\widehat{\bm Y}_{i,m,k}
&=
\bigl\{
\widehat h^{\rm intr},
\widehat p^{\rm intr},
\widehat{\bm h}^{\rm near},
\widehat{\bm p}^{\rm near},
\\
&\qquad
\widehat d^{\rm dist},
\widehat{\bm y}^{\rm intr,aux},
\widehat{\bm Y}^{\rm near,aux}
\bigr\}_{i,m,k}.
\end{aligned}
\label{eq:task_heads}
\end{equation}

Here, $\widehat h^{\rm intr},\widehat{\bm h}^{\rm near}$ denote
conditional first-event hazards;
$\widehat p^{\rm intr},\widehat{\bm p}^{\rm near}$, slice probabilities;
$\widehat d^{\rm dist}$, non-negative distance; and
$\widehat{\bm y}^{\rm intr,aux},\widehat{\bm Y}^{\rm near,aux}$,
intrusion/near-miss descriptors (duration, depth/degree, severity).

\subsection{First-Event Aggregation}
\label{sec:first_event}

For event
$\zeta\in\{\mathrm{intr}\}\cup
\{(\mathrm{near},\rho):\rho\in\mathcal R\}$,
where $\mathcal R=\{R_1,R_2,R_3\}$, let
$y^\zeta_{i,m,k}$ denote its logged-future slice-wise event
indicator. Applying Eqs.~\eqref{eq:predicted_heading}--%
\eqref{eq:predicted_clearance} to the logged-future states
yields the footprints $F^{\rm log}_{i,m,k,j}$ and signed clearances
$d^{\rm foot,log}_{i,m,k,j}$ for $j=1,\ldots,J$.
The event indicators are
\begin{equation}
\begin{gathered}
y^{\rm intr}_{i,m,k}
=
\mathbf 1\!\left[
\min_{1\leq j\leq J}
d^{\rm foot,log}_{i,m,k,j}
\leq \varepsilon_{\rm con}
\right],
\\
y^{\rm near,\rho}_{i,m,k}
=
\mathbf 1\!\left[
\exists j\in\{1,\ldots,J\}:
\varepsilon_{\rm con}
<
d^{\rm foot,log}_{i,m,k,j}
<
\rho
\right].
\end{gathered}
\label{eq:first_event_time}
\end{equation}
The first occurrence slice is
\begin{equation}
T^{\zeta}_{i,m}
=
\min\left\{k : y^{\zeta}_{i,m,k}=1\right\}.
\label{eq:first_occurrence}
\end{equation}
The decoded hazard has the conditional interpretation
\begin{equation}
\widehat h^\zeta_{i,m,k}
=
\Pr\!\left(
T^\zeta_{i,m}=k
\mid
T^\zeta_{i,m}\geq k,\mathcal O_m,\mathcal C_m
\right).
\label{eq:hazard_semantics}
\end{equation}

It induces the unconditional first-event and horizon
probabilities
\begin{align}
\widehat q^\zeta_{i,m,k}
&=
\widehat h^\zeta_{i,m,k}
\prod_{r=0}^{k-1}
\left(1-\widehat h^\zeta_{i,m,r}\right),\\
\widehat P^\zeta_{i,m}
&=
1-\prod_{k=0}^{K-1}
\left(1-\widehat h^\zeta_{i,m,k}\right).
\label{eq:survival}
\end{align}
The decoder also predicts the slice-wise probability that
event $\zeta$ is active:
\begin{equation}
\widehat p^\zeta_{i,m,k}
=
\Pr\!\left(
y^\zeta_{i,m,k}=1
\mid
\mathcal O_m,\mathcal C_m
\right).
\label{eq:slice_event_probability}
\end{equation}

\subsection{Losses and Near-Miss Supervision}
\label{sec:losses}
For stage
$s\in\{\mathrm{dense},\mathrm{entry},\mathrm{calibration}\}$,
the state loss over all valid agents in each candidate-local pool is
\begin{equation}
\mathcal L_{\rm state}^{(s)}
=
\lambda_{xy}^{(s)}\mathcal L_{xy}
+\lambda_v^{(s)}\mathcal L_v
+\lambda_a^{(s)}\mathcal L_a
+\lambda_\omega^{(s)}\mathcal L_\omega
+\lambda_\psi\mathcal L_\psi .
\label{eq:state_loss}
\end{equation}
$\mathcal L_\psi$ is a cosine heading loss.
The risk loss is
\begin{equation}
\begin{aligned}
\mathcal L_{\rm risk}^{(s)}
={}&
\sum_{\zeta}
\Big(
\lambda_{h}^{(s,\zeta)}\mathcal L_{h}^{(\zeta)}
+\lambda_{p}^{(s,\zeta)}\mathcal L_{p}^{(\zeta)}
+\lambda_{B}^{(s,\zeta)}\mathcal L_{B}^{(\zeta)}
\\[-1mm]
&\qquad+
\lambda_{q}^{(s,\zeta)}\mathcal L_{q}^{(\zeta)}
+\bm\lambda_{u}^{{(s,\zeta)}\top}\bm{\mathcal L}_{u}^{(\zeta)}
\Big)
\\[-1mm]
&+
\lambda_d^{(s)}\mathcal L_{\rm dist}
+\lambda_{\rm mono}^{(s)}\mathcal L_{\rm mono}.
\end{aligned}
\label{eq:risk_loss}
\end{equation}
Here, $\mathcal L_h$ and $\mathcal L_p$ are focal-weighted,
class-balanced BCE losses; $\mathcal L_B$ and $\mathcal L_q$ are
Brier losses for horizon occurrence and first-event timing.
$\bm{\mathcal L}_u$ collects masked SmoothL1 losses for duration,
depth/degree, and severity, while $\mathcal L_{\rm dist}$ is a
masked L1 loss on the clipped non-negative minimum
footprint--corridor distance.

The prediction ordering is regularized by
\begin{equation}
\begin{aligned}
\mathcal L_{\rm mono}
&=
\frac{1}{|\Omega_{\rm val}|}
\sum_{(i,m,k)\in\Omega_{\rm val}}
\Big(
[\widehat p^{\rm intr}
-\widehat p^{\rm near,R_1}]_+
\\[-0.3mm]
&\quad+
[\widehat p^{\rm near,R_1}
-\widehat p^{\rm near,R_2}]_+
\\[-0.3mm]
&\quad+
[\widehat p^{\rm near,R_2}
-\widehat p^{\rm near,R_3}]_+
\Big)_{i,m,k}.
\end{aligned}
\label{eq:monotonic_loss}
\end{equation}
where $[z]_+=\max(z,0)$ and $\Omega_{\rm val}$ is the valid-entry set.
Finally,
\begin{equation}
\mathcal L_{\rm total}^{(s)}
=
\mathcal L_{\rm risk}^{(s)}
+\lambda_E\mathcal L_{\rm state}^{(s)}.
\label{eq:total_loss}
\end{equation}

\subsection{Candidate Evaluation and Planner Interface}
\label{sec:candidate_evaluation}

For each event channel $\zeta$, define the worst-agent horizon
risk and early-event urgency as
\begin{equation}
\begin{gathered}
P_m^\zeta
=
\max_{i\in\mathcal I_m}
\sum_{k=0}^{K-1}
\widehat q^\zeta_{i,m,k},
\\
U_m^\zeta
=
\max_{i\in\mathcal I_m}
\sum_{k=0}^{K-1}
\left(1-\frac{k}{K}\right)
\widehat q^\zeta_{i,m,k}.
\end{gathered}
\label{eq:event_selector_terms}
\end{equation}

The intrusion and multi-threshold near-miss scores are
\begin{equation}
\begin{gathered}
R_m^{\rm I}
=
\lambda_{\mathrm P}\bar P_m^{\rm intr}
+\lambda_{\mathrm U}\bar U_m^{\rm intr},
\\
R_m^{\rm N}
=
\sum_{r=1}^{3}w_r
\left(
\lambda_{\mathrm P}\bar P_m^{{\rm near},R_r}
+\lambda_{\mathrm U}\bar U_m^{{\rm near},R_r}
\right).
\end{gathered}
\label{eq:selector_risks}
\end{equation}
where overbar denotes
candidate-wise normalization.

The final candidate score and selection are
\begin{equation}
\begin{gathered}
J_m
=
\frac{
\lambda_{\mathrm I}R_m^{\mathrm I}
+\lambda_{\mathrm N}R_m^{\mathrm N}
}{
\lambda_{\mathrm I}+\lambda_{\mathrm N}}
-\lambda_{\mathrm G}\bar G_m
+\lambda_{\mathrm L}\bar L_m,
\\
m^\star
=
\arg\min_m J_m.
\end{gathered}
\label{eq:candidate_evaluation}
\end{equation}
where $\bar G_m$ and $\bar L_m$ denote the candidate-wise
normalized route progress and absolute final lateral offset,
respectively.

\vspace{2mm}
\begin{algorithm}[H]
\caption{CorrRisk-WM Inference}
\label{alg:corrrisk_wm}
\small
\begin{algorithmic}[1]
\Require Scene $\mathcal{O}$; local pools $\{\mathcal{I}_m\}$;
corridors $\{C_m\}$
\Ensure Risk outputs $\{\widehat{\mathbf{Y}}_{i,m,k}\}$;
selected candidate $m^\star$

\State Encode $z^{E}_{i,m,0},\mu_p,c_{m,k}$ for all $m$
\State Initialize $\hat{x}_{i,m,0}\gets x_{i,m,0}$,
$h^{R}_{i,m,0}\gets h^{R}_{0}$

\For{$k=0,\ldots,K-1$ (all $m$ in parallel)}
    \State Compute $r^{AA}_{i,m,k},r^{AM}_{i,m,k}$
    from $\hat{x}_{i,m,k}$
    \State Evolve $z^{E}_{i,m,k+1}$;
    decode $\hat{x}_{i,m,k,1:J}$
    \State $\hat{x}_{i,m,k+1}\gets\hat{x}_{i,m,k,J}$;
    refresh AA/AM contexts to form $r^{AA}_{i,m,k+1},r^{AM}_{i,m,k+1}$
    \State Compute substep AC geometry; select $\mathcal{A}_{m,k}$
    \State Form $g^{AC}_{i,m,k},r^{AC}_{i,m,k}$;
    update $h^{R}_{i,m,k+1}$
    \State Decode $\widehat{\mathbf{Y}}_{i,m,k}$
    for $i\in\mathcal{A}_{m,k}$
\EndFor

\State Aggregate first-event risks; compute costs $J_m$
\State \Return $\{\widehat{\mathbf{Y}}_{i,m,k}\}$,
$m^\star\gets\arg\min_m J_m$
\end{algorithmic}
\end{algorithm}

\section{Experiments}
\subsection{Dataset and Evaluation Protocol}
\label{sec:dataset_protocol}

We train on 750 Waymo Open Motion Dataset training shards
and evaluate on 100 validation shards ($29{,}176$ scenes)
~\cite{waymo_motion}. All methods share the same candidate
trajectories, IRIS corridors, Top-32 agent pools, deterministic
Top-16 routing, event definitions, evaluator, and planner cost weights.

\subsection{Baselines and Metrics}
\label{sec:baselines_metrics}

\paragraph{Constant Velocity (CV)}
Agents follow
$\bm p_i(t)=\bm p_{i,0}+\bm v_{i,0}t$.
At $0.1$-s substep midpoints, oriented agent footprints use the
cached heading, and their signed distances to the corresponding
corridor polygons are mapped to soft intrusion and
intrusion-suppressed near-miss scores using
$\tau_d=0.25\,\mathrm{m}$.
The maximum substep score defines each slice hazard, and survival
aggregation produces the first-entry probabilities.

\paragraph{Gaussian Overlap-512 (Gauss-512)}
This baseline uses a CV mean with time-growing anisotropic Gaussian
position uncertainty:
\begin{equation}
\bm p_i(t)\sim
\mathcal N\!\left(
\bm p_{i,0}+\bm v_{i,0}t,\,
\bm R_{\theta_i}
\operatorname{diag}\!\left[
\sigma_\parallel^2(t),\sigma_\perp^2(t)
\right]
\bm R_{\theta_i}^{\top}
\right),
\label{eq:gaussian_baseline}
\end{equation}
where $\bm R_{\theta_i}$ is aligned with the velocity direction when
$\|\bm v_{i,0}\|_2>0.5\,\mathrm{m/s}$ and otherwise uses the observed
heading. The standard deviations are
$\sigma_\parallel(t)=0.3\,\mathrm{m}
+(0.4\,\mathrm{m/s})t$ and
$\sigma_\perp(t)=0.2\,\mathrm{m}
+(0.2\,\mathrm{m/s})t$.
At each slice midpoint, 512 samples are drawn using a per-scene seed
and converted to oriented footprints; their intrusion and near-miss
frequencies define slice hazards, followed by survival aggregation.

\paragraph{Latent Dynamics with Geometry (LDG)}
LDG receives the same corridor summaries and uses only state supervision to train its history encoder, AA/AM modules, environment transition, and state probe. its risk is computed analytically from signed clearances between predicted agent footprints and the same time-aligned corridor polygons; negative values indicate overlap.
Intrusion and softly intrusion-suppressed near-miss scores are obtained
from these clearances using sigmoid kernels with
$\tau_d=0.5\,\mathrm{m}$ and thresholds $\rho\in\mathcal R$.
Per-slice hazards use the maximum substep score, followed by the same
survival aggregation as CorrRisk-WM.
LDG contains no learned risk transition or risk decoder.

\paragraph{MLP Geometry (MLP-GEO)}
This non-recurrent baseline maps a 32-D vector of agent motion
and shape, candidate-relative geometry, corridor summaries, and
agent type to intrusion/near-miss hazards and duration/depth
auxiliary outputs using six hidden blocks of width 1280 with
dropout $0.1$.

\paragraph{Metrics}
For intrusion first entry, we report AP, AUROC, Brier score, and
entry MAE, with AP as the primary rare-event metric. Entry MAE measures
probability-weighted entry-time error on intrusion-positive sequences.
Near-miss AP uses intrusion-exclusive first-entry targets for
$\rho\in\mathcal R$. AP, AUROC, and Brier pool all valid
candidate--agent--slice entries. Planner metrics include progress,
selected-corridor intrusion, avoidable intrusion, and raw open-loop
collision; avoidable intrusion requires an intrusion-free alternative.
Raw collision tests the unsmoothed selected self-driving car (SDC) trajectory against all
valid logged agents at synchronized times using a
$5.286\times2.332\,\mathrm{m}$ footprint and is reported as a scene-level rate with a 95\% Wilson confidence interval.


\begin{table*}[t]
\vspace{2mm}
\centering
\caption{Comparison with baseline risk-prediction methods.}
\label{tab:baseline_results}

\begingroup
\normalfont
\fontsize{8}{9.2}\selectfont
\setlength{\tabcolsep}{1.5pt}
\renewcommand{\arraystretch}{0.98}

\begin{tabular*}{\textwidth}{
@{\extracolsep{\fill}}l*{11}{c}@{}
}
\toprule
\multirow{2}{*}{\textbf{Method}}
& \multirow{2}{*}{
    \shortstack{\textbf{Intr.}\\\textbf{AP}$\uparrow$}}
& \multirow{2}{*}{
    \shortstack{\textbf{Intr.}\\\textbf{AUROC}$\uparrow$}}
& \multirow{2}{*}{
    \shortstack{\textbf{Brier}$\downarrow$\\$(10^{-3})$}}
& \multirow{2}{*}{
    \shortstack{\textbf{Entry MAE}$\downarrow$\\(s)}}
& \multicolumn{3}{c}{
    \textbf{Near-miss first-entry AP}$\uparrow$}
& \multirow{2}{*}{
    \shortstack{\textbf{Intr. Proxy}$\downarrow$\\(\%)}}
& \multirow{2}{*}{
    \shortstack{\textbf{Avoid. Intr.}$\downarrow$\\(\%)}}
& \multirow{2}{*}{
    \shortstack{\textbf{Coll.}$\downarrow$\\(\%) [95\% CI]}}
& \multirow{2}{*}{
    \shortstack{\textbf{Progress}$\uparrow$\\(m)}} \\
\cmidrule(lr){6-8}
& & & &
& $0.5\,\mathrm{m}$
& $1.0\,\mathrm{m}$
& $2.0\,\mathrm{m}$
& & & & \\
\midrule

CorrRisk-WM
& 0.8567
& 0.9969
& 2.87
& 0.1788
& \textbf{0.7405}
& \textbf{0.8671}
& \textbf{0.9637}
& \textbf{1.488}
& \textbf{0.199}
& \textbf{4.88} [4.64, 5.13]
& 15.35 \\

MLP-GEO
& 0.7199
& 0.9981
& 1.42
& \textbf{0.0676}
& 0.6064
& 0.7370
& 0.9258
& 1.611
& 0.322
& 5.31 [5.06, 5.58]
& 14.54 \\

LDG
& \textbf{0.8877}
& \textbf{0.9993}
& \textbf{0.45}
& 0.0903
& 0.6632
& 0.7773
& 0.9524
& 1.618
& 0.329
& 7.03 [6.74, 7.33]
& 14.60 \\

CV
& 0.4848
& 0.9433
& 0.81
& 0.2008
& 0.2897
& 0.4400
& 0.8286
& 4.655
& 3.366
& 8.07 [7.76, 8.39]
& 15.89 \\

Gauss-512
& 0.1333
& 0.6061
& 1.14
& 0.4293
& 0.0974
& 0.1611
& 0.6816
& 6.574
& 5.285
& 6.46 [6.18, 6.75]
& \textbf{16.71} \\

\bottomrule
\end{tabular*}
\par\vspace{2pt}

\parbox{\textwidth}{%
\fontsize{7}{8.4}\selectfont
\raggedright
\textit{Protocol:}
The metrics are evaluated using the common risk selector with seed~42.
CorrRisk-WM and LDG use epoch-100 checkpoints, while
MLP-GEO uses its best validation checkpoint.}

\endgroup
\end{table*}


\subsection{Implementation Details}
\label{sec:implementation}

CorrRisk-WM and its controlled ablations use the same internal 5\% split procedure and structured event targets,
with matched training/split seeds $\{42,43,44\}$. Training uses AdamW (peak learning rate $2\times10^{-4}$, weight decay $10^{-4}$), gradient clipping at $1.0$, OneCycleLR with five warm-up epochs, and CUDA AMP for 100 epochs on two NVIDIA H100 GPUs (64 samples/GPU; global batch 128). Training follows a three-stage curriculum $s\in\{\mathrm{dense},\mathrm{entry},\mathrm{calibration}\}$, corresponding to epochs 1--10, 11--90, and 91--100.
MLP-GEO is trained for 50 epochs on an NVIDIA H100 GPU
(batch size 32), and LDG for 100 epochs on two RTX 3090 GPUs (16 samples/GPU; global batch 32).

\noindent\textit{Configuration.}
We set $M=16$, $K=4$, $J=5$, $\Delta=0.5\,\mathrm{s}$,
$\delta=0.1\,\mathrm{s}$, and $H_T=2.0\,\mathrm{s}$.
The input and model dimensions are $N=32$, $N_{\rm sel}=16$,
$P=64$, $L_{\max}=20$, $d=256$, and $N_{\rm hist}=11$, with
$\mathcal A_{\rm type}=\{1,\ldots,8\}$.
We use $\mathcal B=[-10,35]\,\mathrm{m}\times[-10,10]\,\mathrm{m}$
and $(R_1,R_2,R_3)=(0.5,1.0,2.0)\,\mathrm{m}$.
The geometric parameters are $b_\delta=5$,
$v_{\min}=0.1\,\mathrm{m/s}$, $T_{\max}=10\,\mathrm{s}$,
$\beta_{\rm app}=0.35$, $\beta_{\rm ttc}=0.25$,
$\tau_{\rm ttc}=2\,\mathrm{s}$, $\beta_{\rm prox}=1$, and
$\rho_{\rm sel}=0.5\,\mathrm{m}$.
We set $\varepsilon_{\rm norm}=10^{-6}$,
$\varepsilon_{\rm close}=10^{-3}\,\mathrm{m}$, $\varepsilon_{\rm app}=10^{-6}\,\mathrm{m^2}$ and $\varepsilon_{\rm con}=10^{-6}\,\mathrm{m}$.
The selector uses
$(\lambda_{\rm I},\lambda_{\rm N},\lambda_{\rm G},\lambda_{\rm L})
=(0.55,0.30,0.10,0.05)$,
$(\lambda_{\rm P},\lambda_{\rm U})=(2/3,1/3)$, and
$(w_1,w_2,w_3)=(0.5,0.3,0.2)$.

\begin{table}[t]
\caption{Training details of CorrRisk-WM.}
\label{tab:training_details}
\centering

\begingroup
\normalfont
\fontsize{8}{9.2}\selectfont
\setlength{\tabcolsep}{3pt}
\renewcommand{\arraystretch}{1.0}
\setlength{\tabcolsep}{3pt}
\setlength{\aboverulesep}{0.3ex}
\setlength{\belowrulesep}{0.3ex}

\begin{tabularx}{\columnwidth}{
@{}l>{\raggedright\arraybackslash}X@{}
}
\toprule

$E_{\rm E}$
&
State MLP $(8,256,256,256)$;
history projection $(8,256,256)$;
1-layer, 4-head Transformer;
type embedding $(8,256)$
\\

$E_{\rm M}$
&
Point MLP $(4,256,256)$;
masked mean pooling;
polyline embedding;
$2\times$ FFN $(256,512,256)$
\\

$E_{\rm C}$
&
Base MLP $(15,256,256,256)$;
temporal embedding;
optional polygon MLP $(4,256,256)$;
$2\times$ 8-head self-attention
\\

AA/AM
& 8-D geometry descriptors
$\phi^{\mathrm{AA}},\phi^{\mathrm{AM}}$;
bias MLPs $b_{\mathrm{AA}},b_{\mathrm{AM}}$:
$(8,128,1)$ each;
one 8-head AA self-attention and
one 8-head AM cross-attention \\

$T_{\rm E}$
&
Projection $(768,256)$;
GRUCell $(256)$;
residual LN
\\

$D_X$
& State projection $(8,256)$; MLP $(768,512,45)$;
$J=5$ substeps with $9$ outputs each\\

AC
&
Geometry embedding $\bm g^{\rm AC}$ via
$E_g:(10,256)$;
gate/value MLPs
$G_{\rm AC},V_{\rm AC}:(1280,256,256)$ each
\\

$T_{\rm R}$
&
Projection $(1792,256)$;
GRUCell $(256)$;
residual LN
\\

$D_Y$
&
Independent hazard/slice heads $(1536,256,128,o)$;
independent distance/auxiliary heads $(1536,128,o)$
\\

\bottomrule
\end{tabularx}

\vspace{1pt}
\parbox{\columnwidth}{%
\fontsize{6.2}{7}\selectfont
Attention FFNs: $(256,1024,256)$.
Head output: $o=1$ for scalar intrusion, distance, and
auxiliary heads; $o=3$ for near-miss heads.
}

\vspace{2pt}

\fontsize{7.0}{7.5}\selectfont
\setlength{\tabcolsep}{2pt}
\renewcommand{\arraystretch}{1.2}
\setlength{\aboverulesep}{0.45ex}
\setlength{\belowrulesep}{0.45ex}

\begin{tabular*}{\columnwidth}{
@{\extracolsep{\fill}}lccc@{}
}
\toprule
\textbf{Coefficient}
& \textbf{Dense}
& \textbf{Entry}
& \textbf{Calib.} \\
\midrule

$(\lambda_{xy},\lambda_v,\lambda_a,\lambda_\omega)$
& $(.10,.05,.03,.03)$
& $(.05,.03,.02,.02)$
& $(.02,.02,.01,.01)$ \\

\midrule

$(\lambda_h^{\rm intr},\lambda_p^{\rm intr})$
& $(.50,.05)$
& $(1.25,.05)$
& $(.75,.03)$ \\

$\bm\lambda_h^{\rm near}$
& $(.35,.85,.25)$
& $(.30,.75,.20)$
& $(.20,.50,.15)$ \\

$\bm\lambda_p^{\rm near}$
& $(.40,1.00,.25)$
& $(.25,.60,.15)$
& $(.15,.35,.10)$ \\

$\lambda_B=\lambda_q$
& $.05$
& $.05$
& $.20$ \\

$\bm\lambda_u^{\rm intr}$
& $(.15,.15,.10)$
& $(.08,.08,.08)$
& $(.05,.05,.05)$ \\

$\bm\lambda_u^{\rm near}$
& $(.08,.15,.08)$
& $(.04,.08,.04)$
& $(.03,.05,.03)$ \\

$(\lambda_d,\lambda_{\rm mono})$
& $(.05,.05)$
& $(.05,.05)$
& $(.03,.10)$ \\

\bottomrule
\end{tabular*}

\vspace{1.0pt}
\parbox{\columnwidth}{%
\fontsize{6}{6.8}\selectfont
Near-miss vectors follow $(R_1,R_2,R_3)$.
$\lambda_\psi=.05$ and $\lambda_E=1.0$.
}

\endgroup
\end{table}

\subsection{Main Results}
\label{sec:main_results}

Table~\ref{tab:baseline_results} shows that CorrRisk-WM achieves the
best near-miss first-entry AP at all three thresholds
($0.7405$, $0.8671$, $0.9637$), and the lowest selected-corridor
intrusion, avoidable-intrusion, and open-loop collision rates
($0.01488$, $0.00199$, and $0.0488$; 95\% CI:
$[0.0464,0.0513]$). Compared with MLP-GEO, it improves intrusion AP
by $0.1368$ and $1$-m near-miss AP by $0.1301$, reduces collision by approximately $8.1\%$, and increases progress from $14.54$ to $15.35$~m.

The gains are metric-specific. LDG obtains the highest intrusion AP
and AUROC ($0.8877$, $0.9993$) and the lowest Brier score
($0.00045$), while MLP-GEO has the lowest Entry MAE
($0.0676$~s). This is expected because LDG maps predicted
footprint--corridor clearance directly to risk, matching the geometric
construction of intrusion labels. CorrRisk-WM instead provides stronger
multi-threshold near-miss prediction and safer candidate selection,
supporting its role as a planning-oriented risk model rather than a
uniformly dominant predictor on every individual metric.

Selected-corridor intrusion measures logged-agent footprint entry
into the selected corridor slice, whereas open-loop collision
checks SDC--agent footprint overlap along the unsmoothed selected
trajectory. IRIS expands centerline-anchored slices without
enforcing full swept-SDC footprint containment; these related
but distinct metrics are therefore reported separately.

\subsection{Ablation Study}

\begin{table*}[t]
\vspace{2mm}
\centering
\caption{Three-seed ablation study of CorrRisk-WM.}
\label{tab:ablation}

\begingroup
\normalfont
\fontsize{8}{9.2}\selectfont
\setlength{\tabcolsep}{2pt}
\renewcommand{\arraystretch}{0.80}

\newcommand{\abstat}[2]{#1\,\ensuremath{\pm}\,#2}
\newcommand{\abest}[2]{\abstat{\textbf{#1}}{#2}}

\begin{tabular*}{\textwidth}{
@{\extracolsep{\fill}}l*{5}{c}@{}}
\toprule
\textbf{Variant}
& \textbf{Intr. AP$\uparrow$}
& \textbf{Intr. AUROC$\uparrow$}
& \multicolumn{3}{c}{\textbf{Near-miss first-entry AP$\uparrow$}}
\\[-0.8mm]

\cmidrule(lr){4-6}

&
&
&
$0.5\,\mathrm{m}$
& $1.0\,\mathrm{m}$
& $2.0\,\mathrm{m}$
\\[-0.5mm]

\midrule

CorrRisk-WM
& \abest{0.8590}{0.0029}
& \abstat{0.9969}{0.0002}
& \abest{0.7428}{0.0020}
& \abest{0.8705}{0.0031}
& \abest{0.9644}{0.0007} \\

w/o DEM
& \abstat{0.7624}{0.0017}
& \abstat{0.9887}{0.0001}
& \abstat{0.6507}{0.0024}
& \abstat{0.7811}{0.0016}
& \abstat{0.9316}{0.0003} \\

w/o RTRM
& \abstat{0.8530}{0.0026}
& \abest{0.9972}{0.0002}
& \abstat{0.7372}{0.0032}
& \abstat{0.8648}{0.0027}
& \abstat{0.9628}{0.0016} \\

w/o CCGIM
& \abstat{0.7252}{0.0084}
& \abstat{0.9961}{0.0001}
& \abstat{0.5859}{0.0036}
& \abstat{0.7640}{0.0038}
& \abstat{0.9017}{0.0020} \\

\bottomrule
\end{tabular*}

\vspace{3pt}

\begin{tabular*}{\textwidth}{
@{\extracolsep{\fill}}l*{6}{c}@{}}
\toprule
\textbf{Variant}
& \textbf{Brier ($\times10^{-3}$)$\downarrow$}
& \textbf{Entry MAE (s)$\downarrow$}
& \textbf{Intr. Proxy (\%)$\downarrow$}
& \textbf{Avoid. Intr. (\%)$\downarrow$}
& \textbf{Coll. (\%)$\downarrow$}
& \textbf{Progress (m)$\uparrow$} \\
\midrule

CorrRisk-WM
& \abest{2.85}{0.02}
& \abest{0.1772}{0.0014}
& \abstat{1.478}{0.008}
& \abstat{0.190}{0.008}
& \abest{4.86}{0.06}
& \abstat{15.42}{0.10} \\

w/o DEM
& \abstat{3.87}{0.01}
& \abstat{0.2216}{0.0004}
& \abstat{1.698}{0.026}
& \abstat{0.409}{0.026}
& \abstat{5.65}{0.13}
& \abstat{14.60}{0.06} \\

w/o RTRM
& \abstat{3.02}{0.01}
& \abstat{0.1812}{0.0003}
& \abest{1.476}{0.011}
& \abest{0.187}{0.011}
& \abstat{4.90}{0.16}
& \abest{15.46}{0.03} \\

w/o CCGIM
& \abstat{3.84}{0.05}
& \abstat{0.2163}{0.0026}
& \abstat{1.665}{0.014}
& \abstat{0.376}{0.014}
& \abstat{5.15}{0.11}
& \abstat{14.08}{0.12} \\

\bottomrule
\end{tabular*}

\vspace{2pt}

\parbox{\textwidth}{%
\fontsize{7}{8.4}\selectfont
\raggedright
\textit{Protocol:}
Values are mean$\pm$standard deviation over seeds $\{42,43,44\}$;
all runs use epoch-100 checkpoints same training and evaluation protocol.}

\endgroup
\end{table*}
\paragraph{Controlled interventions}
\emph{w/o RTRM} removes $T_{\rm R}$ and zeros the recurrent risk-state
input. \emph{w/o DEM} fixes
$\widehat{\bm x}_{i,m,k,j}=\bm x_{i, m, 0}$ and removes environment-latent
and AA/AM evidence. \emph{w/o CCGIM} keeps the Top-16 router but zeros
$\bm g^{\rm AC}_{i,m,k}$ and $\bm r^{\rm AC}_{i,m,k}$ before risk
updating and decoding.

\paragraph{Observed effects}
Across three seeds, removing DEM reduces intrusion AP from
$0.8590\pm0.0029$ to $0.7624\pm0.0017$, lowers all near-miss APs, and increases the raw collision rate from
$(4.86 \pm 0.06)\%$ to $(5.65 \pm 0.13)\%$.
Removing CCGIM gives the largest intrusion-AP drop
($0.7252\pm0.0084$) and the largest progress loss, confirming DEM and
CCGIM as the main contributors. Removing RTRM causes smaller degradation
in AP, Brier, Entry MAE, and near-miss AP, while downstream selection
metrics remain comparable; RTRM therefore mainly refines risk prediction
and timing.

\section{Discussion and Limitations}
CorrRisk-WM recursively predicts  agent futures using latent and physical states and evaluates their interactions with candidate ego corridors. Rather than reconstructing full-scene sensory futures, it models planning-relevant variables, using each corridor as a spatiotemporal query for intrusion occurrence and timing.

The model remains partial and open-loop: agent evolution is not
conditioned on alternative ego actions, excluding counterfactual
reactions and game-theoretic interactions. Candidate-local filtering
may yield different contexts for the same agent across candidates
despite shared parameters and logged-future supervision.
The transition assumes constant acceleration and yaw rate within
each $0.5$-s step, while the state probe predicts a deterministic
$2$-s future. Footprint geometry and reachability rules provide
conservative priors, not formal safety guarantees. The limited
candidate pool may constrain performance; expanding it increases
corridor-construction and risk-evaluation costs and may increase
planning latency. The selector could also be refined to better
exploit structured risk predictions.

\section{Conclusion}

We presented CorrRisk-WM, which evolves compact scene dynamics and evaluates candidate corridors through future geometric interaction. The model combines latent environment evolution, explicit agent--corridor geometry, and structured risk decoding to estimate intrusion, near-miss, and first-entry risk. Ablations identify dynamic environment modeling and candidate-conditioned interaction as the primary contributors, with recurrent risk modeling providing smaller gains in timing
and probability estimation. Future work will extend the framework to multimodal and counterfactual agent responses, longer horizons, 
and closed-loop collision evaluation, and further improvements in corridor generation, candidate quality,
and trajectory selection.

\appendix
\section{Additional Qualitative Examples}

\begin{figure*}[!t]
    \centering
    \includegraphics[width=\textwidth]
    {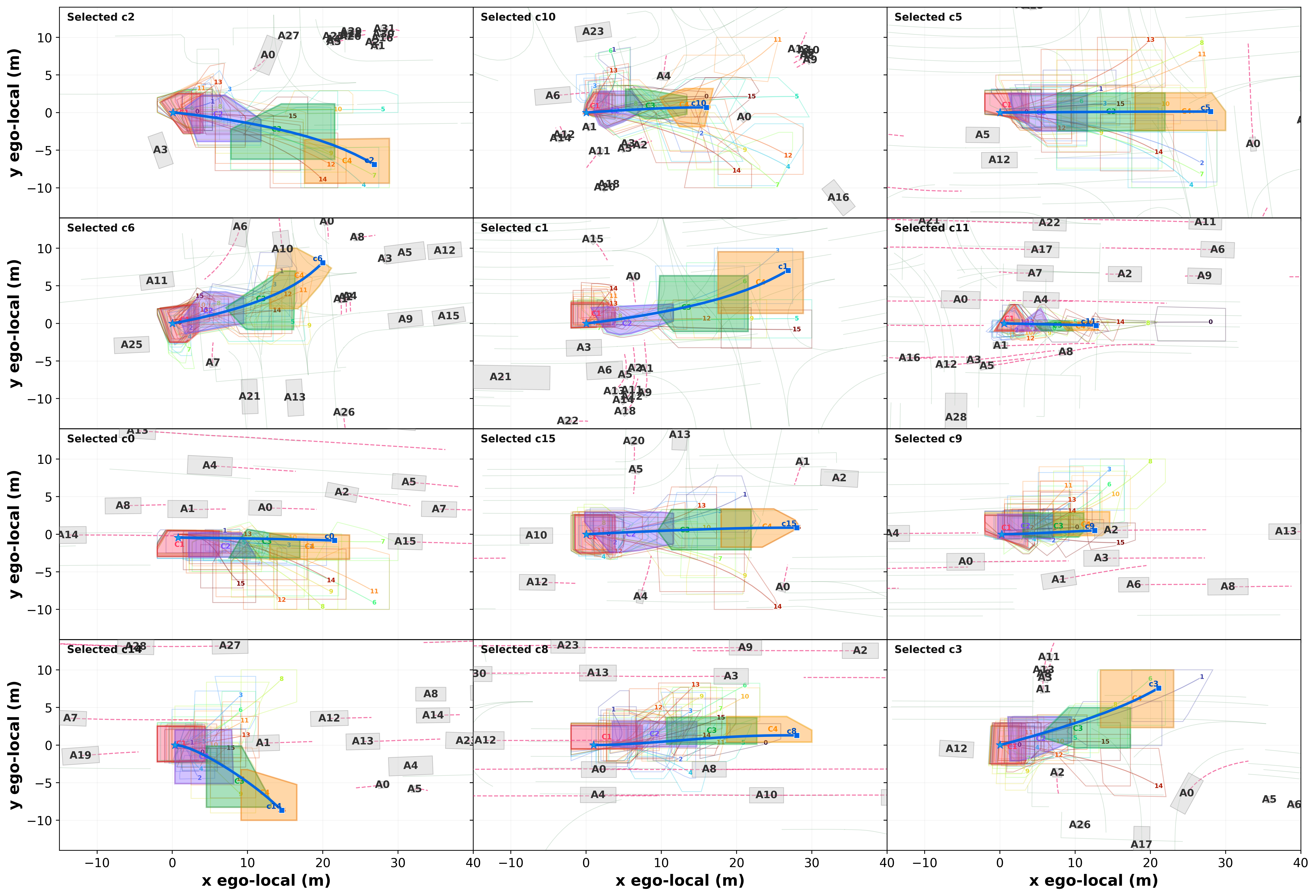}
    \caption{Visualization of corridor-conditioned ego-trajectory selection across 12 Waymo validation scenarios.}
    \label{fig:appendix_ego_candidates}
\end{figure*}

Fig.~\ref{fig:appendix_ego_candidates} presents additional qualitative
examples from the same 100-shard Waymo validation cache used for
quantitative evaluation. For each scene, the upstream sampler provides
$M=16$ candidate ego trajectories and their time-aligned corridor
slices. The trajectory minimizing $J_m$ under CorrRisk-WM predictions is highlighted in blue.

\section*{Acknowledgment}
We used ChatGPT and Codex (OpenAI) to assist with code development, language polishing, and refinement of Fig.~2.

\bibliographystyle{IEEEtran}
\bibliography{references}

\end{document}